\documentclass{article}

\PassOptionsToPackage{numbers, compress}{natbib}
\usepackage[preprint]{neurips_2024}

\usepackage[utf8]{inputenc} % allow utf-8 input
\usepackage[T1]{fontenc}    % use 8-bit T1 fonts
\usepackage[backref=page]{hyperref}
\usepackage{url}            % simple URL typesetting
\usepackage{booktabs}       % professional-quality tables
\usepackage{amsfonts}       % blackboard math symbols
\usepackage{nicefrac}       % compact symbols for 1/2, etc.
\usepackage{microtype}      % microtypography
\usepackage{xcolor}         % colors
\usepackage{wrapfig}

\usepackage{multirow}
\usepackage{amsmath}
\usepackage{amssymb}
\usepackage{cleveref}
\usepackage{graphicx} 
\usepackage{array}

\newcommand{\ours}{Real2Code} 
\usepackage[utf8]{inputenc}

\crefname{section}{Sec.}{Secs.}
\crefname{figure}{Fig.}{Figs.}
\crefname{table}{Tab.}{Tabs.} 

\title{Real2Code: Reconstruct Articulated Objects \\ via Code Generation}

\author{%
  Zhao Mandi \\
  Stanford University\\
  \And
  Yijia Weng \\
  Stanford University\\
  \And
  Dominik Bauer \\
  Columbia University\\
  \And
  Shuran Song \\
  Stanford University\\
}

\begin{document}

\maketitle

\begin{abstract}
  We present Real2Code, a novel approach to reconstructing articulated objects via code generation. Given visual observations of an object, we first reconstruct its part geometry using an image segmentation model and a shape completion model. We then represent the object parts with oriented bounding boxes, which are input to a fine-tuned large language model (LLM) to predict joint articulation as code. By leveraging pre-trained vision and language models, our approach scales elegantly with the number of articulated parts, and generalizes from synthetic training data to real world objects in unstructured environments. Experimental results demonstrate that Real2Code significantly outperforms previous state-of-the-art in reconstruction accuracy, and is the first approach to extrapolate beyond objects' structural complexity in the training set, and reconstructs objects with up to 10 articulated parts. When incorporated with a stereo reconstruction model, Real2Code also generalizes to real world objects from a handful of multi-view RGB images, without the need for depth or camera information. 
  \footnote{Project Website: \url{https://real2code.github.io}}
\end{abstract}

% \footnote{}
% \footnote{Correspondence to: {\tt\small mandi@stanford.edu}}

\begin{figure}[h]
    \centering
    \includegraphics[width=0.98\textwidth]{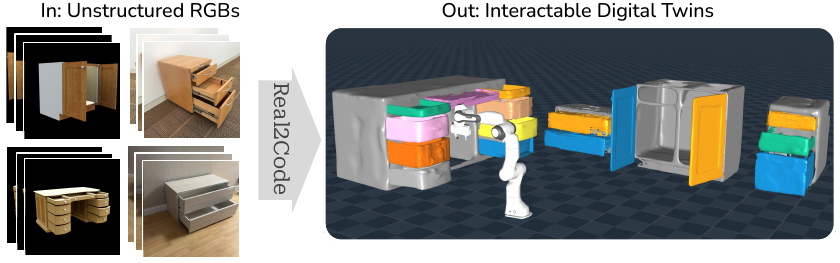}
    \caption{
We propose a novel method for reconstructing articulated objects via code generation, leveraging pre-trained large language models (LLMs). \ours~takes visual observations as input, and performs both part-level geometry reconstruction and joint prediction. When evaluated on an extensive set of real and synthetic objects with varying level of kinematic complexity, \ours~can reconstruct complex articulated objects with up to 10 parts, and generalize to real world objects from a handful of pose-free RGB images.  
    % link: https://docs.google.com/drawings/d/11fRWXrJxqkz-EGtR1m_gyu1rdFY-siOQ7j6nFhhR4VA/edit?usp=sharing 
    }     
    \label{fig:teaser}
    \vspace{-2mm} 
\end{figure}

\section{Introduction}
\label{sec:intro}
The ability to reconstruct real-world objects in simulation (real-to-sim) promises various downstream applications: automating asset creation for building VR/AR experiences, enabling embodied agents to verify their interaction in simulation before execution in the real world~\cite{lim2022planar,real2sim2realgrasping,torne2024reconciling}, or building large-scale simulation environments that support data-driven policy learning~\cite{gen2sim}. We are particularly interested in articulated objects, for both their ubiquity in household and industrial settings and the unique challenges they pose in contrast to single-body rigid objects. 
% \shuran{I feel the following sentences are not very necessary? it is obvious?} The first of these challenges is geometry, i.e., segmenting and reconstructing object parts from partial observations; second is articulation, i.e.,estimating the correct kinematic relationships of object parts (e.g. how a cabinet is connected to its door). 
To reconstruct articulated objects, prior learning-based methods typically train supervised~\cite{ditto} or test-time-optimized~\cite{liu2023paris} models on synthetic objects with simple articulation structures (i.e., one or two moving parts per object). This results in limited generalization ability to objects with more complex visual appearances and kinematics. 
% In addition, prior work does not produce simulation-ready outputs: the extracted object meshes are often incomplete and unstable to simulate, and the predicted articulation parameters require manual cleanup before being used by simulation. 
In addition, prior work only provides object part reconstructions of limited quality: the extracted meshes are often incomplete and the predicted articulation parameters require manual cleanup before being usable for simulation. 
 
We propose \textbf{\ours}, a novel approach to address the above limitations. We represent object articulation with code programs, and use language modeling to predict these code programs from visual observations. This formulation scales elegantly with objects' structural complexity: to process an articulated object with multiple joints, prior methods would require either changing the output dimension of their articulation prediction model, or run multiple inferences on pairs of before- and after- interaction observations to predict one joint at a time. In contrast, the next-token prediction formulation in language modeling allows generating arbitrary-length outputs, i.e., the model architecture needs no adjustment to handle varying number of object joints. Whereas prior work on shape programs~\cite{shapeprog} needs to define task-specific code syntax, we represent objects with simulation code in Python, which takes advantage of recent progress in large language models (LLMs) that are pre-trained with code generation capabilities.

Although capable at code generation, LLMs pre-trained on text are not as equipped at predicting accurate numerical values from spatial geometry information, which is required in our task in order to obtain articulated joint parameters. To address this, we propose to use oriented bounding boxes (OBBs) as an abstraction layer that summarizes the raw sensory observation to the LLM in a concise yet precise manner. Given partial observations of an object, we first perform part-level segmentation and reconstruction via a combination of 2D segmentation and a 3D shape completion model; next, OBBs are extracted from the reconstructed object parts, and serve as input to the LLM. Instead of having to regress to continuous values, the LLMs can predict joints as a classification problem by selecting the closest OBB rotation axis and box edges. 
% We next address part segmentation and reconstruction, which is challenging due to partial observation caused by object occlusion. We propose to use a 2D Segment-Anything \cite{sam} model for its strong generalization ability to in-the-wild data, and fine-tune the model on a custom dataset such that the outputs are adjusted to accurately match an object's kinematic structure, i.e., a partition happens only when two parts are connected by a joint. Then, we group 2D segmentation results into 3D segmented point-clouds, and train a shape-completion model to predict dense occupancy field. The occupancy predictions are used for extracting part-level meshes, effectively completing the partial observation inputs. 

In unstructured real world environments, another challenge is the lack of accurate depth and camera information. To address this, we incorporate a pre-trained dense stereo reconstruction model, namely DUSt3R\cite{wang2023dust3r}, into our pipeline: we show the dense 2D-to-3D point-map prediction from DUSt3R can be combined with our fine-tuned SAM model to achieve view-consistent 3D segmentation. As a result, \ours~can then reconstruct real world objects from only a handful of pose-free RGB images.
% into our pipeline, and with out-of-distribution visual appearances and can be used to animate articulated objects from real scanned static scenes. 
 
For more systematic evaluation, we validate \ours~on the well-established PartNet-Mobility dataset~\cite{partnet}, using an extensive test set of unseen objects that contain various numbers of articulated parts. Compared to the prior state of the art, \ours~significantly improves both 3D reconstruction and joint prediction accuracy. \ours~is the only approach to reliably reconstruct objects with more than three articulated parts, whereas prior methods fail completely on such objects. \cref{fig:teaser} highlights our results on both synthetic multi-part objects (left column of input images), where we show \ours~can reconstruct objects with up to 10 articulated parts, and generalize to real world objects (right column of input images) using RGB images captured from in-the-wild furniture objects.

In summary, our contributions are threefold: 

1. We present \ours, a novel approach to reconstructing articulated objects from a handful of unstructured RGB images. We formulate joint prediction as a code generation problem and adapt pre-trained large language models to specialize in this task.

2. We address part reconstruction via kinematic-aware view-consistent image segmentation and a learned 3D shape completion model, which leads to high-quality mesh extraction that generalizes to multi-part real-world objects. 

3. Empirical results demonstrate that \ours~significantly outperforms the prior state of the art at both articulation estimation and part reconstruction. To the best of our knowledge, \ours~is the first method to accurately predict objects with more than three parts, and generalizes {beyond} the training dataset to objects with up to 10 articulated parts.  
% \begin{enumerate} 
%    \item  We present \ours, a novel approach to reconstructing articulated objects from multi-view RGB-D images. We formulate joint prediction as a code generation problem and adapt pre-trained large language models to specialize in this task.
    
%     \vspace{1mm}
%     \item We address part reconstruction via kinematic-aware view-consistent segmentation and a learned 3D shape completion model, which leads to high-quality mesh extraction that generalizes well to multi-part real-world objects. 
    
%     \vspace{1mm}
%     \item Empirical results demonstrate that \ours~significantly outperforms the prior state of the art at both articulation estimation and part reconstruction. To the best of our knowledge, \ours~is the first method to accurately predict objects with more than three parts, and generalizes {beyond} the training dataset to objects with up to 15 articulated parts.  
% \end{enumerate}

\section{Related Work} 
\textbf{LLMs for Visual Tasks.}
Pre-trained LLMs have been used for visual reasoning and grounding tasks\cite{zeng2022socratic,hsu2023ns3d}. LLMs' code-generation capability has also been exploited for generating programs that solve visual tasks~\cite{visprog, surís2023vipergpt,subramanian2023modular}. These works use zero-shot pre-trained LLMs such as GPTs~\cite{gpt3,openai2023gpt4} and require prompt engineering, such as providing in-context examples, to guide the model to generate desirable outputs; in contrast, we directly fine-tune the weights of a code-generation model to specialize in our articulation prediction task, and do not use any hand-crafted prompting. 

\textbf{Shape Programs.} 
Code-like programs have been studied in computer vision as a compact representation for 2D and 3D shapes. A main challenge for learning code programs is the lack of supervision, and prior work has explored using learned differentiable code executor~\cite{shapeprog}, pseudo-labeling~\cite{jones2022plad}, differentiable rendering~\cite{3d-differentiable}, imitation learning on code sequences~\cite{fusion360}, or reinforcement learning~\cite{Tulsiani2016LearningSA}. More recent work has explored constructing large-scale datasets of shapes\cite{ganin2021computeraided} or scene layouts\cite{avetisyan2024scenescript} and train supervised LLM-like models to generate code outputs. In contrast to ours, these prior work focuses on either individual object shapes or scene-level room layouts, but does not estimate joint articulations. In addition, instead of the task-specific code programs, such as customarily-designed language syntax~\cite{shapeprog,jones2022plad,avetisyan2024scenescript} or Computer-Aided Design (CAD) code~\cite{fusion360,ganin2021computeraided}, we represent object articulation with Python code that 1) closes matches the pre-training distribution of code-generation LLMs, which allows fine-tuning with limited data, and 2) can be directly executed by a physics simulator~\cite{mujoco}, which makes the reconstruction more usable for and requires less manual cleanup.

\textbf{Articulation Model Estimation.}
Prior work has investigated estimating pose and joint properties of articulated objects \textit{without} full reconstruction. A common setup is to assume physical interactions on an object to infer its articulation information: classical sampling-based algorithms\cite{Procrustes-Lo-RANSAC,katz2013interactive} are proposed to estimate joint parameters based on sensory inputs from an object's different configuration states; other learning-based methods train end-to-end models to predict part-level segmentation, kinematic structure, object part poses, or articulated joint parameters~\cite{hu2017learning,yi2018deep, wang2019shape2motion, Michel2015PoseEO, li2020categorylevel, zeng2021visual, huang2021multibodysync,Tseng2022CLANeRF,abdul2022learning,jiang2022opd,liu2023building}. Some propose specialized neural network architectures to improve estimation performance\cite{Buchanan2023OnlineFactorgraph, Heppert22factorgraph, Sun2023OPDFormer}. Other work focuses on learning to propose the most informative physical interactions on an object to help robot manipulation\cite{mo2021where2act}, or to better isolate and segment articulated parts and joints\cite{atp}. These articulation estimation tasks provide useful metrics for 3D shape reasoning \cite{wang2019shape2motion}, and the predicted object pose and joint information are shown useful for robot task learning~\cite{akbNet,An2023RGBManip,geng2023gapartnet, geng2023partmanip}. However, prior work typically handles objects with simple structure (i.e.,one or two moving parts) and does not address full object reconstruction. In contrast, our method handles objects with more than ten moving parts, and performs shape reconstruction via extracting part meshes. 

\textbf{Articulated Object Reconstruction.}
The most closely related to ours are methods that reconstruct both the geometry and joints of articulated objects. A popular approach is to train end-to-end models on synthetic data to jointly segment articulated parts and predict joint parameters, assuming either observations from interactions~\cite{ditto, dittohouse, sfa, mu2021asdf} or single-stage~\cite{Heppert2023carto,kawana2022unsupervised,wei2022self} observations. Another approach uses per-object optimization~\cite{liu2023paris,liu2023building} without training. Based on observations of the object at two or more different joint states, it optimizes for joint parameters to match observed motion correspondence and optionally performs 3D reconstruction by extracting from learned neural rendering fields. Most existing methods assume a single joint and do not scale well with number of joints: for example, to handle an object with $N$ joints, methods like Ditto~\cite{ditto} would need to move the $N$ joints one by one, record the observations before and after each interaction, and run $N$ inferences on each observation pair. PARIS~\cite{liu2023paris} would need to optimize $N$ neural fields and joint parameters, which may lead to a much more complex optimization landscape. The approach presented in~\cite{liu2023building} handles multiple joints but requires a complete sequence of point-cloud observations and is not able to reconstruct 3D shapes.

\section{Method}
We address the problem of reconstructing multi-part articulated objects from visual observations. An articulated object is composed of a set of rigid-body parts that are connected via joints. We assume joint types are either prismatic or revolute: a prismatic joint is parameterized by a joint axis $\mathbf{u}^p \in \mathbb{R}^3$ and a translation offset $d$; a revolute joint is parameterized by a position $\textbf{p}^r \in \mathbb{R}$, a rotation axis $\mathbf{u}^r \in \mathbb{R}^3$, and a rotation angle $\theta$. For an object with $N$ moving parts, we assume each to be connected with its parent via exactly one 1-DoF joint. Therefore, the transformation between each part's frame and its parent's frame is uniquely determined by the joint parameters --- this observation motivates our OBB-based formulation described in \cref{method-code}. 
% : $^n_{n-1}T(\mathbf{u}^p, d)$ for prismatic joints and $^n_{n-1}T(\mathbf{u}^r,\textbf{p}^r, \theta)$ for revolute joints. 
To obtain visual input of our system, we assume an object is manipulated such that each of its articulated joints is at a non-zero state, i.e.,$d>0$ or $\theta >0$, and record a set of RGB (and optionally depth) images of the object. Our system outputs a set of 3D meshes -- each a reconstruction of the object's parts -- and a list of joint types and parameters represented as code. The meshes and joints can then be used to create the object's digital twin in simulation and enable downstream applications.

\begin{figure}[t]
    \centering    \includegraphics[width=0.98\linewidth]{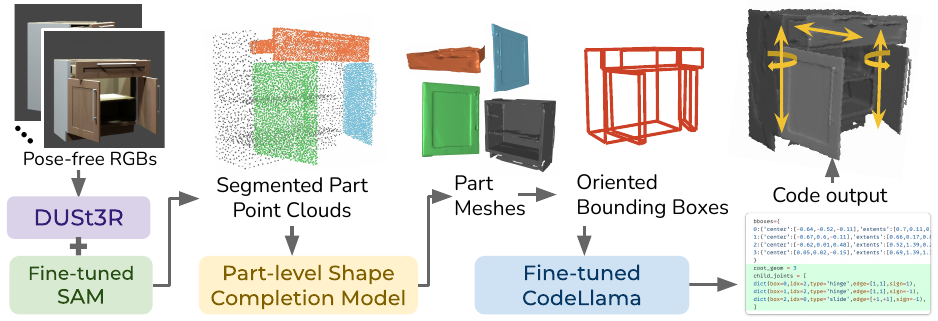} 
    \caption{Overview of our proposed pipeline. Given unstructured multi-view RGB images, we leverage the pre-trained DUSt3R model~\cite{wang2023dust3r} to obtain dense 2D-to-3D pointmaps, and use a fine-tuned 2D segmentation model\cite{sam} to perform part-level segmentation and project to segmented 3D point clouds. A learned shape-completion model takes partial point cloud inputs and predicts a dense occupancy field, which is used for part-level mesh extraction. We fine-tune a large language model (LLM)~\cite{codellama} that takes mesh information in the form of oriented bounding boxes, and outputs full code descriptions of the object that can directly be executed in simulation.
    }
    \label{fig:method}
    \vspace{-5mm}
\end{figure}

\cref{fig:method} provides an overview of our proposed method. \ours~consists of two main steps: reconstruction of object parts' geometry (\cref{method-shape}) and joint estimation via LLM code generation (\cref{method-code}). Between the two steps, the oriented bounding boxes (OBBs) of the object parts serve as an abstraction layer, enabling the LLM to reason about 3D spatial information and predict accurate joint parameters. We introduce the two main steps in the following two sections.

\subsection{Part Reconstruction}\label{method-shape}
To reconstruct an object's part-level shape geometry, we propose a 2D-to-3D approach that is category-agnostic and able to address objects with arbitrary number of parts. First, we fine-tune a SAM model that generates 2D segmentations from RGB images, and projects them to 3D point clouds. Next, we train a shape completion model that extracts watertight meshes from the partially-observed point clouds. 
% We describe the method details below.
\vspace{-4mm}

\subsubsection{Kinematics-aware Part Segmentation} 
\vspace{-1mm}
We leverage a pre-trained 2D segmentation model to first segment object parts based on their kinematic structure. This design is motivated by the need to 1) generalize to real world data, and 2) scalably segment arbitrary number of object parts. In contrast to prior works that train 3D segmentation models with limited synthetic data~\cite{ditto, partnet, sapien}, the SAM~\cite{sam} model was pre-trained on a much larger scale dataset and hence generalizes well to in-the-wild real world images. Further, in contrast to prior works that infer articulation from multiple object states, we leverage SAM's strong prior to identify object parts without the need for multi-step interactions. 

However, because SAM~\cite{sam}~is originally designed for iterative user prompting, its zero-shot predictions display uncontrollable granularity on articulated objects, i.e.,segmenting unnecessary details that require additional user input to refine. To address this, we propose to adapt the pre-trained SAM using the PartNet-Mobility \cite{partnet,sapien} dataset: while keeping the model's heavy-weight image encoder frozen, we fine-tune the lightweight prompt-decoder layer of SAM to take an image and one sampled 2D point prompt as input, and predict the corresponding mask that matches the object's kinematic structure. More details on the fine-tuning process are reported in \cref{app:training}. 

% We fine-tune the lightweight prompt-decoder layer of SAM while keeping the larger image encoder frozen. We use $28,020$ training RGB images where each image corresponds to the same number of segmentation masks as object parts plus a background mask. A size $B$ training batch contains $B$ RGB images, and $B\times16$ pairs of GT masks and 2D query points sampled for each image. We update the model with the same loss as proposed in the original paper~\cite{sam}, i.e., a weighted average of Focal Loss~\cite{lin2018focal}, Dice Loss~\cite{diceloss} and MSE IoU prediction loss.

\subsubsection{Test-time Prompting for View-consistent Segmentation.} %Due to the point-prompt-based formulation, the model can predict any many object parts, but also has the downside of being inherently semantics unaware.
\begin{wrapfigure}{r}{0.598\linewidth}
% \begin{figure}
    \centering   
    % \vspace{-5mm}
    \includegraphics[width=0.95\linewidth]{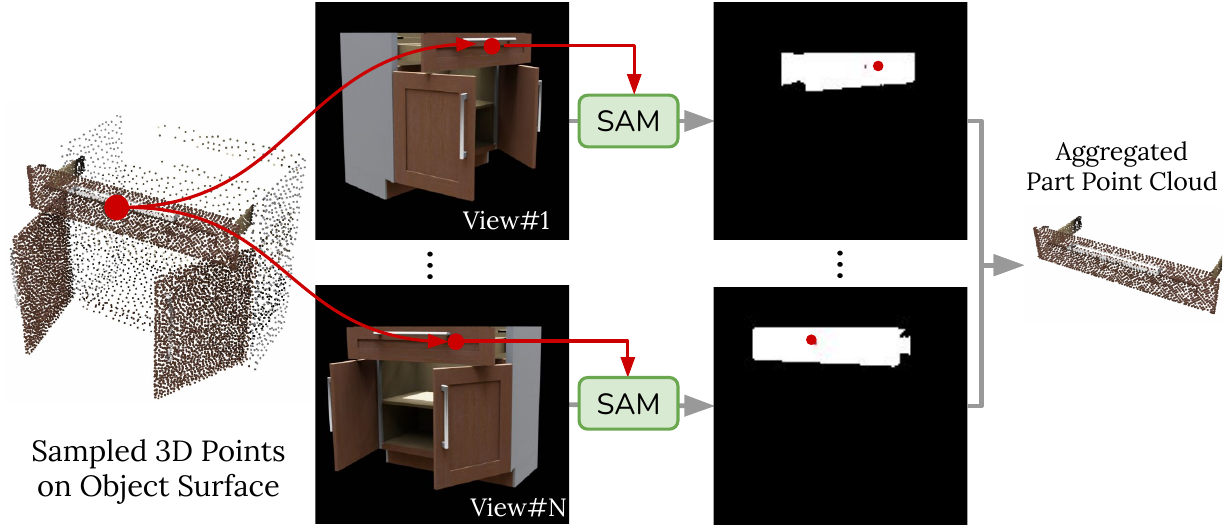}
    \caption{\footnotesize \textbf{View-consistent segmentation.} Illustration of our method for test-time prompting the fine-tuned SAM model. We first sample 3D points from the foreground object point clouds, project each point onto 2D RGB images captured from different camera views, which are used to prompt the model to generate view-consistent segmentations.}
    % link to figure (feel free to make edits & update here!): https://docs.google.com/drawings/d/1kbYRY8yaZ8Luz88qQvJ_QNkoM4YS24AO_THyS6VNdjE/edit
    % \vspace{-7mm}
% \end{figure} 
\end{wrapfigure}
The point-based segmentation described above allows our fine-tuned SAM to scale easily with the number of the object parts. However, this formulation also inherently lacks view consistency, as SAM is unaware of the object part correspondences across different camera views. To address this, we introduce a test-time prompting procedure to project predicted 2D masks into a view-consistent 3D segmentation. We discuss two different input settings based on the availability of depth and camera data: \textbf{1) Multi-view RGB-D and Camera Input}: we first coarsely sample 2D points on each RGB image and run the SAM model to obtain the background masks. This allows us to segment the foreground object in the different views and sample 3D points uniformly on the point cloud. Next, we project each such 3D point back onto each image, and obtain view-consistent 2D points for SAM prompting. Further, we rank the model predicted masks based on the confidence and stability scores proposed in~\cite{sam}, and filter them using non-maximum suppression~(NMS) to produce the final 3D segmentation. \textbf{2) Multi-view Unstructured RGB Input.} To handle real world settings which often lack high-quality depth and camera information, we adopt a multi-view stereo reconstruction model to achieve part segmentation from unstructured, pose-free RGB images. We use the recently proposed DUSt3R\cite{wang2023dust3r} model, which is pre-trained to predict dense 3D point-maps from RGB input images. We then sample 2D points from one RGB image and find each point's corresponding point in every other RGB images via nearest-neighbor. More details are described in \cref{app:realobj}. This overall procedure of projecting between 3D to 2D prompting is similar to SA3D~\cite{cen2023segmentnerf}, which samples on a NeRF\cite{mildenhall2020nerf} field and uses inverse rendering to effectively prompt SAM in 3D.
\vspace{-2mm}
\subsubsection{Part-level Shape Completion.}
Due to frequent self-occlusion in articulated objects, e.g. the inside of a drawer is often not visible, RGB-D input does not provide full observation of each object part, and subsequently a segmented point cloud does not recover complete shape geometries. This motivates learning a shape completion model to obtain watertight meshes. Because part-segmentation is already handled in the previous step, we here tackle shape completion on the object part level. We build on top of Convolutional Occupancy Nets~\cite{convocc}: the model architecture consists of a PointNet++\cite{pointnetpp} point-cloud encoder, followed by a 3D Unet~\cite{unet3d} encoder and a linear MLP decoder that predicts logits for occupancy. We use the ground-truth part meshes from PartNet-Mobility\cite{partnet} to generate a dataset of partial point cloud inputs and occupancy labels. We normalize the occupancy grid using \textit{partial} OBBs extracted from the input point cloud to avoid under-fitting the smaller-sized meshes. Marching Cubes~\cite{marchingcube} is used to extract the completed part meshes from predicted occupancy. See \cref{app:training} for more details on our shape completion dataset and model training.

% The model is trained using a binary cross entropy loss using the ground-truth occupancy data generated from the PartNet-Mobility~\cite{partnet,sapien} assets. 
% We use $6,260$ pairs of partial point clouds and size $96^3$ occupancy grids and train our PointNet++~\cite{pointnetpp} based occupancy prediction model from scratch. For a training batch of size $B$, we sample $B$ point clouds of size $2048$, and sample $B\times12,000$ query points on the label occupancy grids. Notably, because object parts are of different scales, we normalize the occupancy grid using \textit{partial} OBBs extracted from the input point cloud to avoid under-fitting the smaller-sized meshes. When sampling training query points, we found sampling 25\% occupied works the best for balancing between occupied areas and empty space, and we add a random shifting step on the occupied grids to improve model accuracy on the near-surface areas. At test time, we query on a $96^3$ grid and use Marching Cubes~\cite{marchingcube} to extract the completed part meshes. 
\vspace{-2mm}
\subsection{Articulation Prediction via Code Generation}
\label{method-code} 
Given a set of segmented object parts, the next step is to predict the articulation structure that connects the parts. Our approach of using LLM code generation offers several advantages: first, code is a compact representation for articulation, and when combined with LLM's ability to predict arbitrary-length outputs, it scales elegantly with the complexity of an object's kinematic structure; second, pre-trained LLMs are equipped with strong priors for both common-sense objects and for generating syntactically correct code, making them easily adaptable to our task; lastly, the LLM-generated code can be directly executed in simulation, removing the need for manual cleanup of predicted joint parameters, which is required by prior work~\cite{ditto}. The following sub-sections first introduce our formulation of joint parameters w.r.t. OBBs, then discuss our proposed fine-tuning procedure that adapts a pre-trained LLM to our articulation prediction task. 

\subsubsection{Oriented Bounding Box as Input Abstraction.}
\begin{figure}[t]
    \centering    \includegraphics[width=0.98\linewidth]{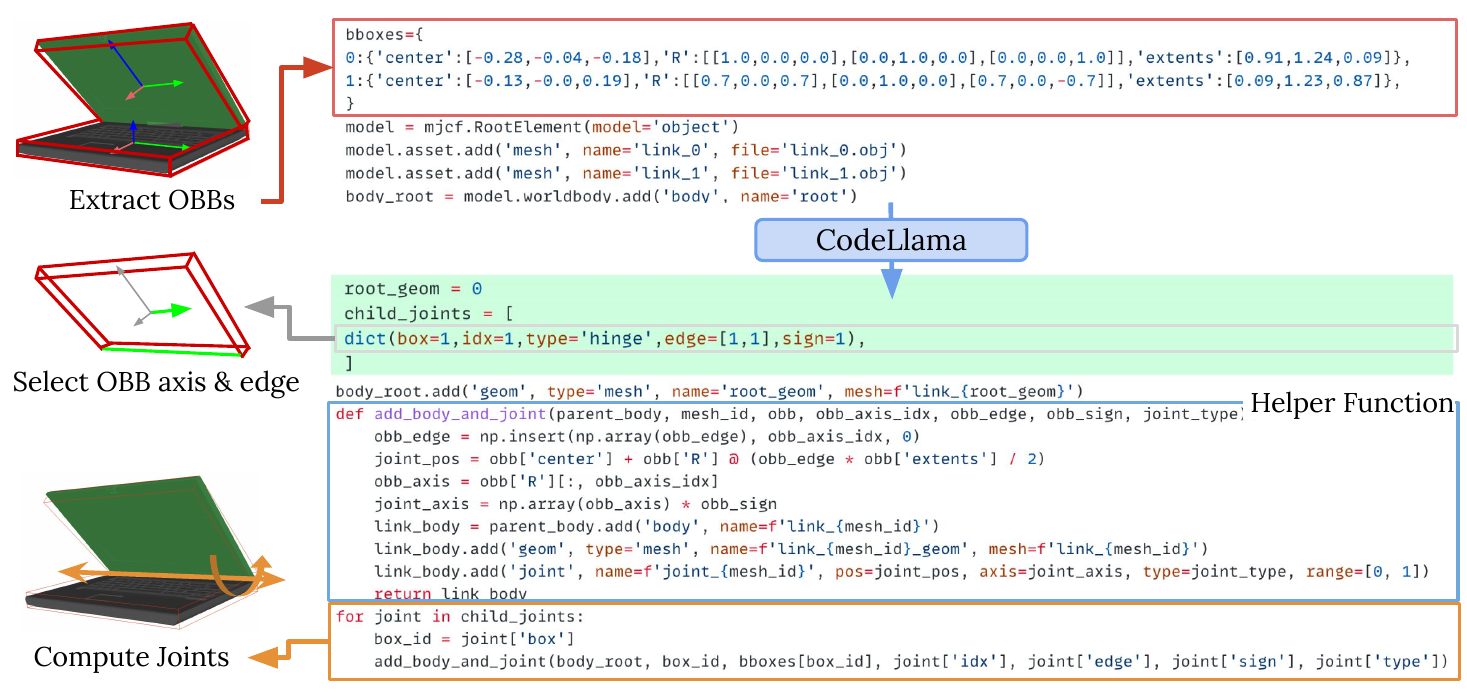}
    \caption{\textbf{Articulation Prediction as Code.} We fine-tune a Codellama model that takes in oriented bounding boxes (OBBs) for segmented parts as input, and generates joint predictions via selecting OBB rotation axes and edges (model generation is highlighted in green). A helper function is used to compute the absolute joint axis and position that assembles the object parts in simulation}
    % link to code-method figure (feel free to edit!): https://docs.google.com/drawings/d/1ZR4RSzGJSIiaFZkqqxaTBOeCImycsM90259dzZdei1I/edit
    \label{fig:method-code} 
\end{figure}

Articulation prediction requires numerical precision at joint parameters (i.e., position and rotation) and reasoning from raw sensory input, but an LLM pre-trained on text is not naturally adept at these challenges. We address this by representing the sensory input (object point clouds) as a set of oriented bounding boxes (OBBs), each representing a segmented and completed object part. Compared to alternative object representations such as 3D point clouds or 2D images, OBBs strike a balance between \textbf{compactness} (i.e., do not require an extra feature encoder) and \textbf{preciseness} (i.e., provide numerical 3D pose information). Further, OBBs provide a reference for joint information. Recall that the pose of an object part is determined by its 1-DoF joint at a non-zero state --- we can hence recover joint parameters from the observed displacement of object parts. Given an OBB of a part connected to its parent, the joint axis will be parallel to one of the three axes of the OBB's rotation matrix regardless of its joint type. We also observe many common articulated objects consist of cuboid-like parts (e.g. doors or laptops), hence the position of their corresponding revolute joints will lie close to one of the OBB edges. Combining these observations, we can re-formulate the joint axis prediction problem by selecting an OBB rotation axis as the joint axis and, for revolute joints, choosing an OBB edge parallel to the axis as the joint position. See~\cref{fig:method-code} for an illustration. 

\subsubsection{Fine-tuning a Code Generation LLM.}
We now have an input formulation that effectively converts a regression task (i.e., predicting continuous values) to an easier classification task (i.e., selecting axes and edges) for LLMs. We use the 7B-CodeLlama~\cite{codellama} model for its open-source-availability and built-in priors for code generation. We construct a fine-tuning dataset using PartNet~\cite{partnet} objects (the same asset used to generate our segmentation and shape completion data above), where the native URDF files are converted into MJCF code~\cite{dmcontrol} that 1)~is in the more compact Python syntax, 2) can be executed in MuJoCo~\cite{mujoco} physics simulation, and 3) has each object's joints assigned with respect to the corresponding part's OBB information. The LLM takes a list of part-OBB information (i.e., center, rotation, and half-lengths) as input, and outputs joint predictions as a list, where each line contains indices into the axes and edges of an OBB. More details on LLM fine-tuning can be found in \cref{app:training}

\section{Experiments}
\vspace{-2mm}
We evaluate \ours~and compare to baseline methods to validate the effectiveness of our approach. \cref{exp-seg} describes experiments on our kinematics-aware 2D image segmentation and 3D shape completion models. \cref{exp-joint} evaluates our fine-tuned code-generation model on articulation prediction. \cref{exp-ablate} contains ablation studies that provide additional insights into our method. In \cref{exp-real}, we test our trained models on real world object data and qualitatively demonstrate the generalization ability of \ours. 
\vspace{-1mm}
\subsection{Experiment Setup}
\vspace{-1mm}
\textbf{Datasets.} We use assets from five categories in PartNet-Mobility\cite{partnet} dataset: Laptop, Box, Refrigerator, Storage-Furniture and Table. The same split of 467 train and 35 test objects are used to construct our image segmentation, shape completion, and code datasets. We use Blender~\cite{blender,blenderproc} to render RGB-D and ground-truth segmentation masks for each object. The RGB-D images and masks are then used to generate part-level point clouds as partial observations. For code data, we extract OBBs from part meshes and process each object's raw URDF file into Python MJCF\cite{dmcontrol} code, where the rotation and position of each joint are relative to the OBB of the child part that this joint connects to the parent part. Refer to \cref{app:dataset} for more details on our dataset construction.
% The objects are split once into training and testing objects, such that the same set of objects are used for training all three models and the other unseen objects are used for testing.

\textbf{Baselines.} We compare \ours~to the following baseline methods:

$\bullet$ \ \ \textbf{PARIS}~\cite{liu2023paris} is the prior state-of-the-art for articulated object reconstruction. It takes multi-view RGB observations of a two-part articulated object at two different joint states, then optimizes per-part NeRF-based reconstructions and joint parameters based on motion cues from the two observed states. We render our test objects at two random joint states, run their proposed optimization procedure with 5 random initialization seeds, and report the average performance. To handle more complex objects, we modify their method to optimize for more than two parts at once. However, we observe that their design of optimizing one neural field for each part runs out of memory when the number of joints exceeds 4.

$\bullet$ \ \  \textbf{Ditto}~\cite{ditto} is an end-to-end learning method that takes in a pair of before- and after-interaction point cloud inputs and predicts implicit part shape reconstructions and joint parameters. Notably, Ditto assumes only one object part is moved at a time, which requires step-by-step interactions and observations, making evaluation less efficient. For an object with $N$ joints, we move one part at a time, render the corresponding $N$ pairs of point cloud observations, and run their pre-trained model $N$ times to obtain joint parameter predictions and reconstructions of moved parts.
    
$\bullet$ \ \  \textbf{GPT-4}~\cite{openai2023gpt4} is representative of recent state-of-the-art LLMs with strong reasoning and code-generation capability. We use it as a reference for zero-shot LLM performance on our desired task without fine-tuning. We prompt it with the same code header used in our LLM fine-tuning dataset, plus additional instructions on how to format the output, which our fine-tuned LLM does not need.

 \begin{table}
\centering
\small 
% \begin{tabular}{r|cccccccccc}
\begin{tabular}{m{2.4cm}|m{0.6cm}m{0.6cm}m{0.6cm}m{0.6cm}m{0.6cm}m{0.6cm}m{0.6cm}m{0.6cm}m{0.6cm}m{0.6cm}}
\hline
  % & \multicolumn{3}{c}{Laptop (5)} & \multicolumn{3}{c}{Box(2)} & \multicolumn{3}{c}{Refrigerator (5)} & \multicolumn{2}{c}{\footnotesize Storage (20)} \\
  Category & \multicolumn{2}{c}{Laptop} & \multicolumn{2}{c}{Box} & \multicolumn{2}{c}{Fridge} & \multicolumn{2}{c}{ Furniture} & \multicolumn{2}{c}{Furniture}\\

  Number of Parts & \multicolumn{2}{c}{2} & \multicolumn{2}{c}{2} & \multicolumn{2}{c}{2-3} & \multicolumn{2}{c}{2-4} & \multicolumn{2}{c}{5-15}\\
 % \cline{2-2} 

  Metric & \multicolumn{1}{l}{whole} & \multicolumn{1}{l}{part} & \multicolumn{1}{c}{whole} & \multicolumn{1}{c}{part} &  \multicolumn{1}{c}{whole} & \multicolumn{1}{c}{part}  &   \multicolumn{1}{c}{whole} & \multicolumn{1}{c}{part} &   \multicolumn{1}{c}{whole} & \multicolumn{1}{c}{part} \\ \midrule
\ours+\scriptsize gtSeg & 0.57 & 2.33 &  1.54 & 7.65 & 0.51 & 2.04 &   1.46 & 13.3 & 5.84 & 16.8 \\ \midrule
Ditto &2.54 & \textbf{2.04} &  1.73 & 82.82 &  2.80 &\scriptsize 462.25 &   \textbf{2.25 }& \scriptsize 1105.86 &  \textbf{2.21} & \scriptsize 4608.08 \\  
PARIS & 84.29 &\scriptsize 206.31  &  15.35 & \scriptsize 158.73  &  20.63 & \scriptsize 1297.27   &  6.02 &\scriptsize 544.64 &  11.44 & \scriptsize 816.86\\ \midrule  
% Real2Code & \textbf{4.64} & \textbf{8.81} &  & \textbf{14.25} & \textbf{13.59} &  & \textbf{7.33} & \textbf{7.03} &  & \textbf{1.44} & \textbf{1.58} \\ \hline

\ours-\scriptsize SegOnly & 1.74 & 7.19 &  11.46 & 10.52 &  0.90 & 23.44  & 17.43 & 206.49  & \scriptsize N/A & \scriptsize N/A \\

\ours~\scriptsize(Ours) & \textbf{0.44} & 3.02  & \textbf{1.31} &\textbf{5.94} &   \textbf{0.60} & \textbf{1.28} &   3.47 & \textbf{65.79} &  19.70 &\scriptsize \textbf{118.58}\\
\bottomrule
\end{tabular} 
\vspace{1mm}
\caption{We evaluate surface reconstruction quality by measuring Chamfer-Distance (CD) between predicted and ground-truth meshes. Results are reported separately for each object category, where we take average CD of objects' entire surface reconstruction (`whole' column) and of all part wholes (`part' column). Objects from Storage-Furniture and Table are reported under Furniture and divided based on the number of parts.}
\label{tab-shape}
\vspace{-3mm}
\end{table}

\subsection{Part Segmentation and Reconstruction Experiments}
\label{exp-seg} 

\textbf{3D Part-level Shape Completion.}

Following the prompting procedure described in~\cref{method-shape}, we first run our fine-tuned SAM on images from the test set of 35 unseen objects and obtain segmented part point clouds. We observe that, because we rank and filter the mask predictions (i.e., prioritize high predicted confidence score and stability score), the low-quality masks have less impact on the final segmented point-cloud after the projection step. Next, we use the segmented part point clouds as input to evaluate our learned shape completion model, and use Marching Cubes~\cite{marchingcube} on the occupancy predictions to extract meshes. Following prior work~\cite{mu2021asdf,ditto,liu2023paris}, we uniformly sample $10,000$ points on the extracted mesh surface and report the average Chamfer Distance between the extracted and ground-truth part meshes in \cref{tab-shape}. 
Because the predictions are semantics-agnostic, we generate permutations of the set of predicted meshes and take the permutation that results in lowest error; the same logic is used for joint prediction results. 

We remark on the performance difference between \ours~ and baseline methods: the joint optimization of shapes and joints in PARIS~\cite{liu2023paris} suffers from a complex loss landscape and produces unsatisfactory reconstructions, especially when the number of parts increases. Ditto~\cite{ditto} performs well on training categories (i.e., Laptop) but does not generalize well to unseen categories. In contrast, we factorize the problem into segmentation and shape completion, aggregate 2D segmentation from fine-tuned SAM and perform shape completion in a per-part fashion leads to better results. 

Due to the partial observation and noise in the segmentation masks, simply extracting meshes from the grouped point clouds also results in subpar reconstruction results (see column `Real2Code-SegOnly', where `N/A' indicates the mesh extractions are too noisy to match with GT mesh). This further validates the need for our shape completion model. Overall, \ours~achieves the best reconstruction quality and elegantly scales to a larger number of parts.
% \vspace{-1cm}

\textbf{Kinematics-aware 2D Image Segmentation.}

To demonstrate the effectiveness of SAM fine-tuning, we evaluate the fine-tuned model on unseen object images by uniformly sampling a grid of $32\times32$ query points and compare the predicted segmentation with ground-truth masks. We use NMS filtering on the predicted masks, then by sorting with the model's predicted confidence score to take the top-K masks that fill the image to more than 95\% total pixels. We observe a significant improvement over zero-shot SAM: object parts are segmented much more closely following their kinematics structure, obtaining a $92\%$ mean IoU score on the final used masks and $84\%$ match rate to ground-truth masks. 
% Full details are provided in supplementary material.  
% We report results in Table~\ref{result:sam2d}. 
% \input{tables/exp-sam-2d}
\vspace{-3mm}
\subsection{Articulation Prediction Experiments}
\label{exp-joint}
After completing part-level reconstruction on the test objects, we extract OBBs for each object part and compose a text-prompt for our fine-tuned CodeLlama model. We parse the model's code generation and append it with code header lines (e.g. import packages) such that the post-processed code can be directly executed to produce object simulation. We then evaluate the accuracy of articulation prediction by measuring the error of joint type, joint axis, and (for revolute joints only) joint position predictions. 
% \begin{table}
% \label{result:joint}
% \centering
% \begin{tabular}{cccclccclccc}
% \hline
%  & \multicolumn{3}{c}{2 Parts (19)} &  & \multicolumn{3}{c}{3 Parts (9)} &  & \multicolumn{3}{c}{4-15 Parts (9)} \\ 
%  \cline{2-4} \cline{6-8} \cline{10-12} 
 
% \multicolumn{1}{l}{} & rot$\downarrow$ & pos$\downarrow$ & type$\uparrow$ &  & rot$\downarrow$ & pos$\downarrow$ & type$\uparrow$ &  & rot$\downarrow$ & pos $\downarrow$ & type$\uparrow$ \\ \midrule
% Ditto &  &  &  &  &  &  &  &  &  &  &  \\ \hline 
% PARIS &  &  &  &  &  &  &  &  &  &  &  \\ \hline 
% GPT4 &  &  &  &  &  &  &  &  &  &  &  \\ \hline 
% Real2Code+GT-OBB & 10.0 & 0.06 & 0.95  &   & 0.0 & 0.04 & 1.00  &  & 6.43 & 0.04 & 0.97 \\ \hline
% Real2Code (Ours) &  &  &  &  &  &  &  &  &  &  &  \\ \hline 
% \bottomrule
% \end{tabular}
% \vspace{1mm}
% \caption{We report joint prediction results grouped by the number of moving parts in each object. \ours can predict joints accurately on objects with 4 or more moving parts, whereas baseline methods fail}
% \end{table}

\begin{table}
\small
% \scriptsize
\centering
% \begin{tabular}{r|ccc|ccc|ccc|ccc}
\begin{tabular}{m{2.1cm}|m{0.5cm}m{0.5cm}m{0.5cm}|m{0.5cm}m{0.5cm}m{0.5cm}|m{0.5cm}m{0.5cm}m{0.5cm}|m{0.5cm}m{0.5cm}m{0.5cm}}
\hline
 & \multicolumn{3}{c}{2 Parts (15)}  & \multicolumn{3}{c}{3 Parts (9)}  & \multicolumn{3}{c}{4-5 Parts (6)}   & \multicolumn{3}{c}{6-15 Parts (7)} \\ 
 %\cline{2-4} \cline{6-8} \cline{10-12} \cline{14-16} 
 
 & rot$\downarrow$ & pos$\downarrow$ & type$\uparrow$   & rot$\downarrow$ & pos$\downarrow$ & type$\uparrow$  & rot$\downarrow$ & pos$\downarrow$ & type$\uparrow$  & rot$\downarrow$ & pos$\downarrow$ & type$\uparrow$    \\ \midrule

\text{\raggedleft Real2Code+{\scriptsize gtBB}} & 0.0 & 0.07 & 0.93  & 0.0 & 0.04 & 1.00  & 0.0 & 0.04 & 1.00 & 11.6 & 0.03 & 0.94 \\

\midrule
 
Ditto & 40.04 & 4.04 & 0.57  & 35.57 & 2.47 & 0.70 &  49.77 & 3.20 & 0.43 & 63.06 & 4.16 & 0.30  \\  
PARIS & 48.44 & 2.67 & 0.51  & 32.35 & 3.63 & 0.84 & 55.97 & 2.14 & 0.43 &  N/A  & N/A  & N/A \\  
GPT4 &   57.3 & 0.26 & 0.73  & 10.0 & 0.08 & 0.61  & 45.0 & 0.21 & 0.51 & \textbf{30.0} & \textbf{0.05} & 0.71 \\  
\midrule
\ours~{\scriptsize(Ours)} & \textbf{7.5} & \textbf{0.08} & \textbf{0.80 }&  \textbf{0.0} & \textbf{0.04} & \textbf{0.89} &   \textbf{0.63} & \textbf{0.07} & \textbf{0.97} & 30.2 & \textbf{0.05} & \textbf{0.89} \\ 

\bottomrule  
\end{tabular} 
\vspace{1mm}
\caption{Joint prediction results from \ours~and baseline methods, grouped by the number of moving parts in each object. We remark that \ours~consistently outperforms baseline methods across objects with different kinematic structures; on objects with 4 or more moving parts, \ours~predicts joints accurately whereas baseline methods fail.}
\vspace{-7mm}
\label{result:joint}
\end{table}

% 10.00 & 0.08 & 61.11

% \begin{table*}[]
% \centering
% \begin{tabular}{cccclccclccc}
% \hline
%  & \multicolumn{3}{c}{Laptop, Box (7)} &  & \multicolumn{3}{c}{Refrigerator (5)} &  & \multicolumn{3}{c}{StorageFurniture (20)} \\ \cline{2-4} \cline{6-8} \cline{10-12} 
% \multicolumn{1}{l}{} & \multicolumn{1}{l}{rot} & \multicolumn{1}{l}{pos} & \multicolumn{1}{l}{type} &  & \multicolumn{1}{l}{rot} & \multicolumn{1}{l}{pos} & \multicolumn{1}{l}{type} &  & \multicolumn{1}{l}{rot} & \multicolumn{1}{l}{pos} & \multicolumn{1}{l}{type} \\ \midrule
% Ditto &  &  &  &  &  &  &  &  &  &  &  \\ \hline 
% PARIS &  &  &  &  &  &  &  &  &  &  &  \\ \hline 
% GPT4 &  &  &  &  &  &  &  &  &  &  &  \\ \hline 
% Real2Code (ours) & 0.00 & 0.02 & 85.71  &   & 0.00 & 0.01 & 90.00  &  & 0.00 & 0.01 & 69.24 \\
% \bottomrule
% \end{tabular}
% \end{table*}

As shown in~\cref{result:joint}, we outperform all baseline methods by a large margin. The effectiveness of our OBB abstraction is further accentuated by Real2Code+gtBB, where we feed oracle OBB to the code generation module and achieve highly accurate predictions even on unseen objects with a large number of parts.
  
\subsection{Qualitative Results}
\begin{figure}
    \centering
    \includegraphics[width=0.99\linewidth]{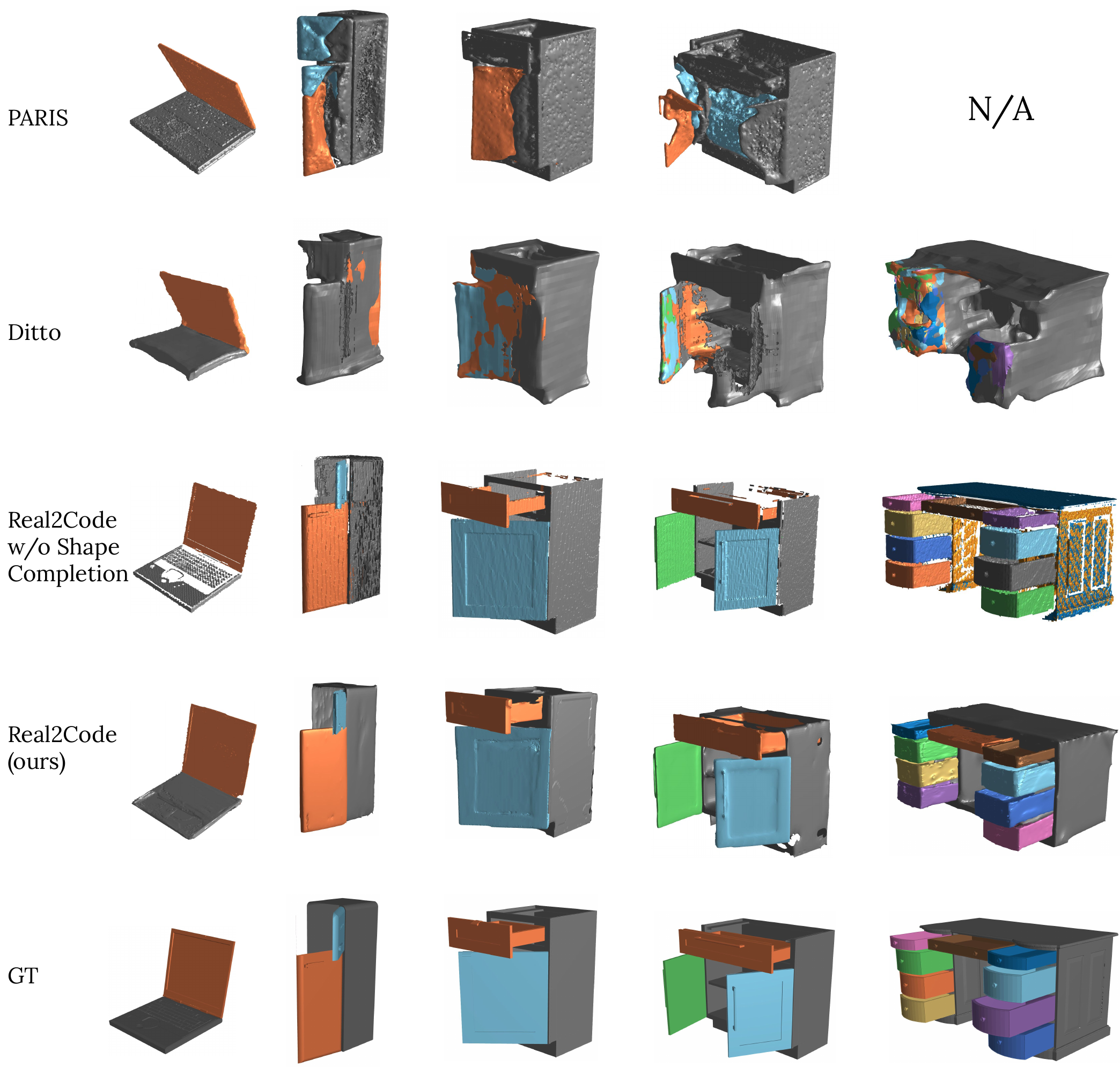}
    \caption{Qualitative results that compare \ours~to baseline methods. We show results from objects with a range of varying kinematic complexities, from a two-part laptop to a ten-part multi-drawer table. Whereas all methods can handle the simpler laptop articulation, baseline methods struggle as the number of object parts increases, and \ours~performs reconstruction much more accurately. PARIS runs out of memory and fails to run on the ten-part table (`N/A').
    % link: https://docs.google.com/drawings/d/15hKgUaOsFM7M6fPhJUq6jLZEOxxStH6PKWFcQN1cXB8/edit?usp=sharing
    % TODO: add joint visualization later
    \vspace{-2mm}
    }
    \label{fig:results-qual}
\end{figure}
For qualitative results, we select objects with a range of varying kinematic complexities, from a two-part laptop to a ten-part multi-drawer table. We visualize the final reconstructed objects from ours and baselines methods in ~\cref{fig:results-qual}. Whereas all methods can handle the simpler laptop articulation, baseline methods struggle as the number of object part increases, and Real2Code performs reconstruction much more accurately.

\subsection{Ablation Studies}
\label{exp-ablate}
To further validate our formulation of using OBB as reference for articulation prediction, we provide additional ablation experiments that use alternative input and output representation:

$\bullet$ \ \ \textbf{Regression on Joint Parameters.} Instead of selecting OBB rotation axis and edge,  we fine-tune two more CodeLlama models to take the same input but outputs continuous numerical values for joint parameters: the first model directly predicts 3 values for each joint axis and 3 for every joint position (\cref{exp-ablate} row `OBB Abs.'); the second model predicts joint axis the same way as~\ours, but predicts joint position as a relative position to the OBB's center (\cref{exp-ablate} column `OBB Rel.').
    
$\bullet$ \ \ \textbf{Provide LLM with Visual Inputs.} To verify whether OBBs provide sufficient spatial information, we fine-tuned another model with both RGB and OBB inputs. We adopt the OpenFlamingo~\cite{alayrac2022flamingo,awadalla2023openflamingo} approach for interleaving image embeddings with the CodeLlama model weights, and uses the same pre-trained ViT~\cite{vit} weights for image encoder.   
\begin{table}
 % \vspace{-3mm} 
\small
\centering

% \begin{tabular}{ll|cccl|cccl|cccl|ccc}
\begin{tabular}{m{0.55cm}m{0.55cm}|m{0.5cm}m{0.5cm}m{0.6cm}|m{0.5cm}m{0.5cm}m{0.6cm}|m{0.5cm}m{0.5cm}m{0.6cm}|m{0.5cm}m{0.5cm}m{0.6cm}}

\hline
 & & \multicolumn{3}{c}{2 Parts (15)}   & \multicolumn{3}{c}{3 Parts (9)}   & \multicolumn{3}{c}{4-5 Parts (6)}  & \multicolumn{3}{c}{6-15 Parts (6)} \\ 
 %\cline{2-4} \cline{6-8} \cline{10-12} \cline{14-16} 
 
Inp. & Out & rot$\downarrow$ & pos$\downarrow$ & type$\uparrow$ &  rot$\downarrow$ & pos$\downarrow$ & type$\uparrow$ &  rot$\downarrow$ & pos$\downarrow$ & type$\uparrow$ &  rot$\downarrow$ & pos$\downarrow$ & type$\uparrow$  \\ \midrule

% OBB & Abs. & 15.9 & 0.23 & 0.87 & 4.0 & 0.11 & 1.00 & 0.0 & 0.08 & 1.00 & 2.8 & 0.28 & 0.89\\  
OBB & Abs. & 7.5 & \textbf{0.06} & 0.92 &  \textbf{0.0} & \textbf{0.03} & \textbf{1.0}   & \textbf{0.0} & 0.6 & 0.83 &  \textbf{0.0} & 0.7 & 0.73 \\  

OBB & Rot. & 0.0 & 0.18 & 0.73 & 0.3 & 0.23 & 1.00 & 0.9 & 0.19 & 0.83 & 5.9 & 0.06 & 0.59 \\

% +RGB & Rel. & 0.5 & 0.27 & 0.80 & 0.3 & 0.03 & 1.00 & 2.0 & 0.04 & 1.00 & 3.8 & 0.03 & 0.96 \\   
+RGB & Cls. & \textbf{0.0} & \textbf{0.06} & 0.80  & 5.0 & \textbf{0.03} & \textbf{1.0}  & \textbf{0.0} & \textbf{0.03} & 0.89 & \textbf{0.0} & \textbf{0.02} & 0.67 \\

% \midrule     
% OBB & Rel. &  15.0 & 0.14 & 0.80 & 20.3 & 0.03 & 1.00 & 2.0 & 0.03 & 1.00 & 14.3 & 0.02 & 0.96\\
\midrule  
OBB & Cls. & \textbf{0.0} & 0.07 & \textbf{0.93}  & \textbf{0.0} & 0.04 & \textbf{1.0}  &  \textbf{0.0} & 0.04 & \textbf{1.00} &  11.6 & 0.03 & \textbf{0.94} \\ 
 
\bottomrule  
\end{tabular} 
\vspace{1mm}
\caption{Joint prediction results from ablation experiments on \ours. Using the `Regress' output formulation, the LLM is still able to output reasonable values for two or three part objects, but generates much less accurate joint positions when the number of articulated parts increase. Adding additional RGB image input yields no clear improvements from the model, which suggests the OBB input alone can provide sufficient information for articulation.}
\vspace{-5mm}
\label{result:ablate-llm}

\end{table}

Results from the ablation experiments are reported in~\cref{result:ablate-llm}. We make the following remarks:

\textbf{Regression formulation predicts less accurate joint positions.} Both predicting absolute joint positions (column `OBB Abs.') and predicting relative position from OBB center (column `OBB Rot.') yield a higher prediction error than selecting OBB edges as joint position. In contrary, the rotation error is still on a reasonable scale: we found this is due to the model has learned to copy the correct axis column from the OBB rotation matrices contained in the input prompt. This further validates the effectiveness of using OBB as spatial representation. 

\textbf{RGB input does not yield significant improvement.}
We draw this conclusion from comparing row `+RGB Rel.' and `OBB Rel.'. This suggests the OBB input provides sufficient information for articulation prediction task. 

% \begin{figure}[t]
%     \centering
%     \includegraphics[width=0.99\linewidth]{figures/real2code-quali-code.pdf}
%     \caption{Qualitative comparison of the code output format in ablation experiments. Each joint prediction occupies one line. In `Regress`, the LLM outputs continuous position and axis values, i.e., six parameters per joint; \ours~ uses `Classify`, where LLM outputs three parameters: one value selects rotation axes in an OBB, and two values to indicate the OBB edge. For `RGB+OBB` ablations, we provide one RGB image as additional input to the LLM; we show an example in the bottom right of the figure.
%     % the helper functions (not shown in the figure) then use the predicted values to compute joint parameters and add corresponding mesh to parent body. 
%     }
%     \label{fig:code-qual}
%     \vspace{-3mm} 
% \end{figure}

% output: regression on joint (global)/ relative to bbox (local)
% code: with/without helper function 
% input: bbox, image 

% segmentation: with/without finetuning 
% reconstruction: Ditto/Paris/Ours 

 \begin{wrapfigure}{r}{0.598\linewidth} 
    \centering    
    \includegraphics[width=0.99\linewidth]{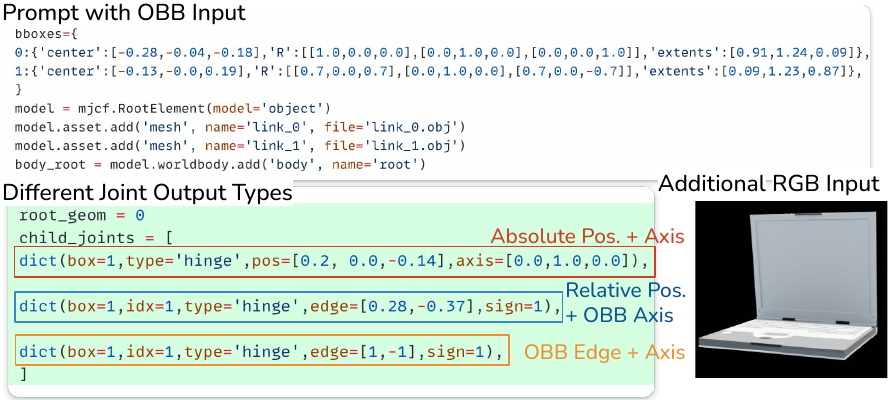}
    \caption{\small Qualitative comparison of the code output format in our ablation experiments. Each prediction format occupies one line. In `Absolute Pos. + Axis', the LLM outputs continuous position and axis values; in `Relative Pos. + OBB Axis', the LLM outputs one index into the OBB's rotation axis, and a 2D joint position relative to the selected axis; \ours~ uses `OBB Edge + Axis`, where LLM outputs index to rotation axes in an OBB, and two values to indicate the OBB edge. Bottom right of the figure shows one example of additional RGB image input to the LLM.}
    % link to figure (feel free to make edits & update here!): https://docs.google.com/drawings/d/1kbYRY8yaZ8Luz88qQvJ_QNkoM4YS24AO_THyS6VNdjE/edit
    \vspace{-3mm}
% \end{figure} 
\end{wrapfigure}

\subsection{Experiments on Real World Objects}
\label{exp-real}
\begin{figure}
    \centering
    \includegraphics[width=0.98\linewidth]{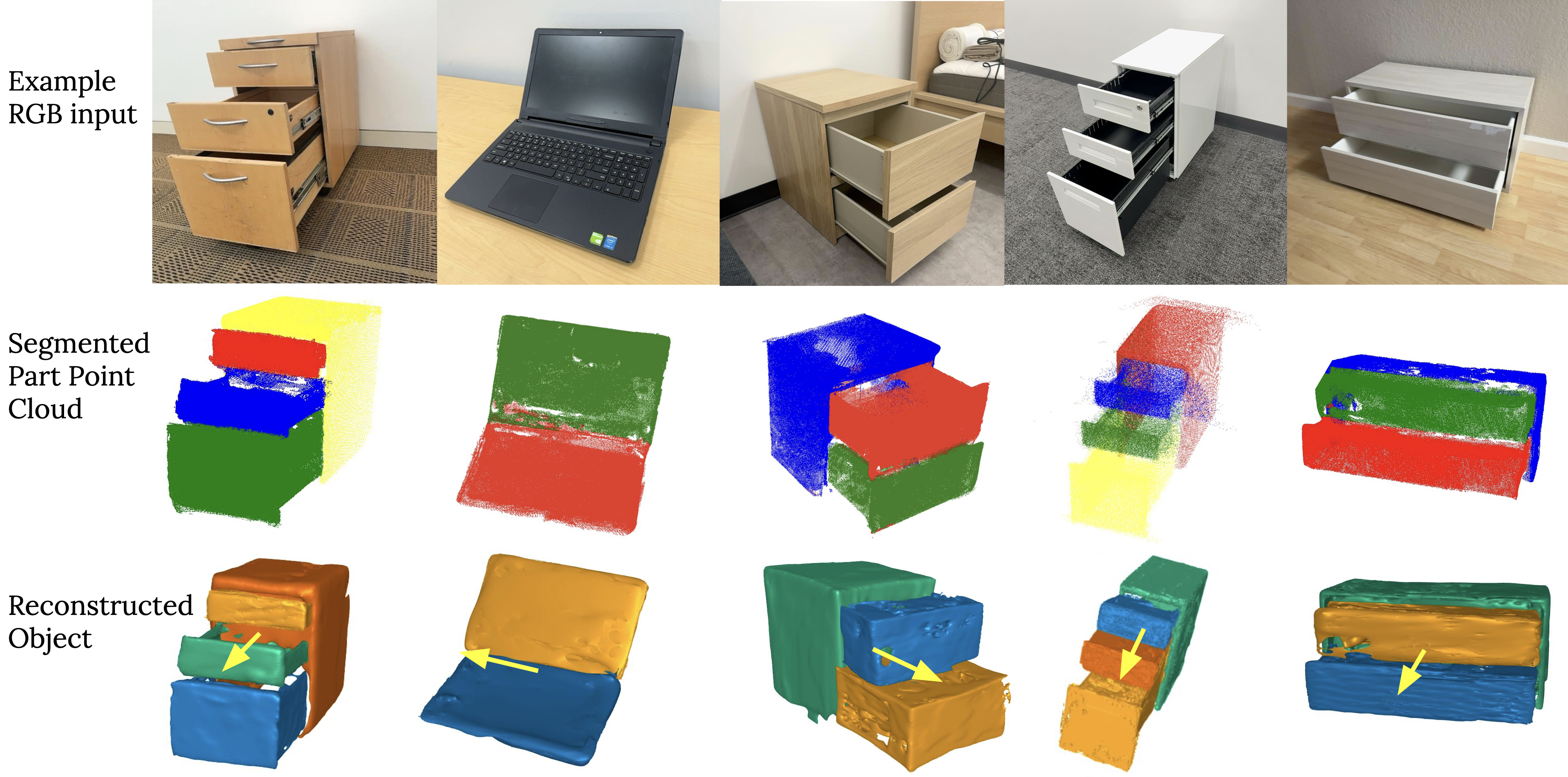}
    \caption{We evaluate \ours~on real world objects using RGB data. For each object, we use 10 pose-free RGB images captured in-the-wild and run \ours~with DUSt3R\cite{wang2023dust3r}. We show one example RGB input (1st row), segmented point clouds (2nd row) and full reconstruction (3rd row) for each object.
    % link: https://docs.google.com/drawings/d/15hKgUaOsFM7M6fPhJUq6jLZEOxxStH6PKWFcQN1cXB8/edit?usp=sharing
    % TODO: add joint visualization later
    }
    \label{fig:results-realobj}
\end{figure}

To validate the generalization ability of \ours, we gather a set of in-the-wild articulated objects and collect multi-view RGB data as inputs. We run \ours~with DUSt3R\cite{wang2023dust3r}. Due to the achieve reconstruction from multi-view pose-free RGB images. Due to the lack of quantitative metrics, we show qualitative results in \cref{fig:results-realobj} that \ours~generalizes well to in-the-wild objects and produces good quality reconstructions from RGB-only inputs. However, although the learned DUSt3R\cite{wang2023dust3r} model performs well on overall shape and exterior surface areas of the objects, it predicts less accurate point maps at areas inside the drawers, which is likely due to the lack of similar data in their training dataset. As a result, the segmented part point clouds display noises (second row in \cref{fig:results-realobj}), which leads too lower quality mesh extraction from the shape completion model. See \cref{app:realobj} for more details on our evaluation setup.

\section{Limitations}
In its current form, \ours~still has a few limitations that point to interesting directions for future work: 
1) \ours~only considers single object with many parts, extending it to multiple object scenes would require additional object detection and preprocessing. 
2) \ours~only predicts joint parameters in terms of their type, position, and axis. To infer other joint parameters, such as joint range and friction world require additional multi-step interactions and observations. 
3) We found that the articulation prediction accuracy is sensitive to failures in the first 2D image segmentation module, i.e., OBBs from wrong segmentations directly obstruct the LLM reasoning of object structures; this can be improved by providing human corrective feedback as proposed in~\cite{sam}, i.e., a user provides additional points and prompt the model to adjust its mask predictions.

\section{Conclusion}
We present~\ours, a novel method for reconstructing articulated objects that leverages code generation capability in pre-trained LLMs. We empirically show that \ours~achieves a new state-of-the-art in both geometry reconstruction and articulation prediction and can successfully reconstruct objects with complex kinematic structures with an arbitrary number of parts, whereas prior methods fail. By reliably translating visual observation to simulatable models, we hope \ours~unlocks new opportunities in robotics and mixed reality applications.

\section*{Acknowledgements}
The authors would like to thank Zhenjia Xu for the real world data collection, Samir Gadre for helpful pointers on LLM fine-tuning, Cheng Chi for help with training our shape completion model training, and all members of REALab: Zeyi Liu, Xiaomeng Xu, Chuer Pan, Huy Ha, Yihuai Gao, Mengda Xu, Austin Patel, et. al. for valuable feedback and discussion on the paper manuscript. We also thank Stanford EE department admins Kenny Green, Steve B. Cousins and Mary K. McMahon for their support during real world data collection. This work was supported in part by the Toyota Research Institute, NSF Award \#2143601, Sloan Fellowship.  The views and conclusions contained herein are those of the authors and should not be interpreted as necessarily representing the official policies, either expressed or implied, of the sponsors. 
                
{
\small

\bibliography{refs}
\bibliographystyle{neurips_2024}
}

%%%%%%%%%%%%%%%%%%%%%%%%%%%%%%%%%%%%%%%%%%%%%%%%%%%%%%%%%%%%

\appendix

\section{Appendix}
\subsection{Dataset Preparation Details}
\label{app:dataset}
\textbf{Base: PartNet-Mobility Object Assets.}
We use the same set of 468 training and 41 testing objects from 4 categories in PartNet-Mobility~\cite{partnet}. The raw dataset contains a rich collection of object meshes, textures, and URDF files that contain articulation information. We further process the data as follows:

 \textbf{RGB-D Image Rendering} We render each object individually using Blender~\cite{blender,blenderproc} for 5 loops. For each rendering loop, the object is centered at the scene origin and the rendering camera poses are randomly sampled; we render 12 RGB-D images and all the segmentation masks corresponding to the all the object parts. During rendering, we also randomly sample joint states in the object such that all its doors or drawers are partially open --- we make the assumption that all the parts our train and test objects are partially open to remove ambiguity and provide more observation view into object insides. 
 
\textbf{Mesh Pre-processing.}
The original PartNet-Mobility assets contain highly fine-grained meshes, i.e.,one drawer part is comprised of more than ten panel or bar-shaped meshes. To prepare data for part-level shape completion, we group these fine-grained meshes such that meshes from the same object part are merged into one single mesh. Mesh textures are ignored during grouping, resulting in grouped texture-less part-level meshes. The RGB-D images and masks are then used to generate part-level point clouds as partial
observations. We use Kaolin~\cite{kaolin} to sample label occupancy values from object part meshes. 

\textbf{Code-Generation Data.}
To prepare data for fine-tuning code-generation LLMs, we first use the rendered RGB-D images and segmentation masks to obtain \textit{ground-truth} part-level point-clouds, which are used to extract oriented-bounding boxes (OBBs) for each part. Next, we take the raw object URDF files and generate a shorter copy with our grouped part meshes. Because the raw URDF/XML syntax contain long unnecessary details, we manually translate them into Python-like MJCF~\cite{dmcontrol} code, which are a lot more compact and familiar to the pre-trained LLMs. Finally, for each of the 5 rendering loops per object, we re-write the object code again to replace the absolute joint information with the relative position and rotation of each joint with respect to the extracted OBBs. We further augment the data by randomly rotating the OBBs along the z-axis, 5 times per object. This results in $468\times 5\times 5 = 11700$ training samples for LLM fine-tuning. 

\subsection{Model Training Details}
\label{app:training}
\textbf{SAM Fine-tuning.} The fine-tuning data consists of $28,020$ RGB images, and each image corresponds to a set of binary segmentation masks, one per each object part plus a background mask. We fine-tune only the decoder layers of pre-trained SAM\cite{sam} on this custom dataset while keeping the rest of the model weights frozen. Each fine-tuning batch contains 24 RGB images; for every RGB image in the batch, we sample 16 prompt points uniformly across each image's ground-truth masks, i.e.,only sample points from the positive mask area. Hence each training batch of size $B=24$ contains 24 images and $24\times 16$ pairs of prompt point and ground-truth masks. Following the original paper~\cite{sam}, we update the model with a weighted average of Focal Loss~\cite{lin2018focal}, Dice Loss~\cite{diceloss} and MSE IoU prediction loss.

\begin{figure}[h]
% \begin{figure}
    \centering   
    % \vspace{-5mm}
    \includegraphics[width=0.95\linewidth]{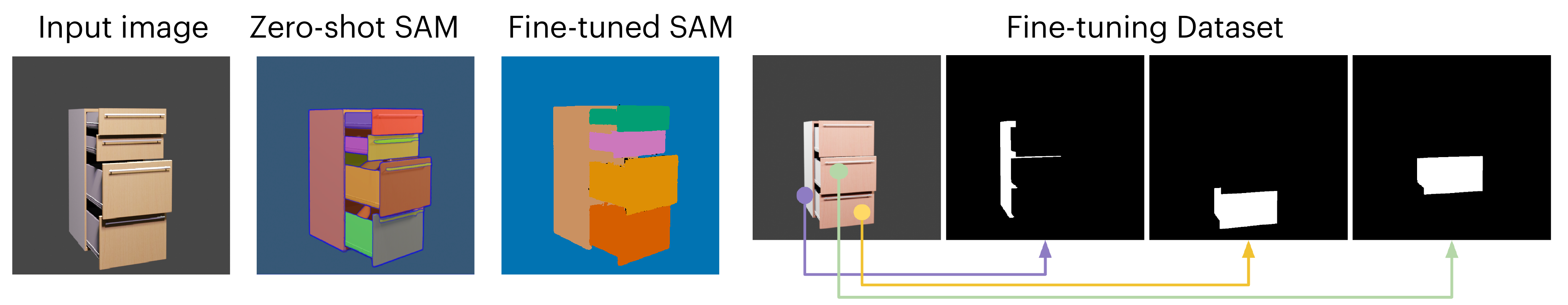}
    \caption{\textbf{Kinematics-aware SAM Fine-tuning.} Given an RGB input image, the pre-trained zero-shot SAM\cite{sam} produces unnecessarily detailed segmentation masks (column ~Zero-shot SAM'. We construct a dataset of objects' RGB images and kinematics-aligned ground-truth masks (column `Fine-tuning Dataset'). The model is fine-tuned to take one image and one sampled 2D query point and predict the corresponding part mask. We compare the output of the model after fine-tuning on the same image (column `Fine-tuned SAM').} 
% \end{figure} 
\end{figure}

\textbf{Training Shape Completion Model.}
We use $6,260$ pairs of partial point clouds and size $96^3$ occupancy grids and train our PointNet++~\cite{pointnetpp} based occupancy prediction model from scratch. For a training batch of size $B$, we sample $B$ point clouds of size $2048$, and sample $B\times12,000$ query points on the label occupancy grids. Notably, because object parts are of different scales, we normalize the occupancy grid using \textit{partial} OBBs extracted from the input point cloud to avoid under-fitting the smaller-sized meshes. When sampling training query points, we found sampling 25\% occupied works the best for balancing between occupied areas and empty space, and we add a random shifting step on the occupied grids to improve model accuracy on the near-surface areas. At test time, we query on a $96^3$ grid and use Marching Cubes~\cite{marchingcube} to extract the completed part meshes. 

\textbf{Fine-tuning Code Generation LLM.}
We use the pre-trained Codellama\cite{codellama}-7B model on our code dataset, which contains code samples generated from PartNet\cite{partnet} objects as described above. We use LoRA~\cite{lora}, a low-rank weight fine-tuning technique, to fine-tune the model with the next-token prediction loss. For training efficiency, we compress the training sequences by removing unnecessary empty character spaces and overhead code lines (such as package import statements). The resulting training set contains under 800 tokens per sequence for objects with up to 7 parts (i.e., 6 articulated joints). Despite the short training data, we found the model to be able to extrapolate to unseen test set objects with up to 15 parts. 

\subsection{Details on Real World Evaluations}
\label{app:realobj}
\textbf{Data Collection.} We collect data from a set of common furniture objects, including cabinets, laptops, night stands, dressers, ranging from 1 to 3 moving parts. Each object is scanned using a LiDAR-equipped iPhone camera and 3dScanner App~\cite{3dscanner} to capture a set of RGB images from the front 180$^{\circ}$ view. We then select 10 RGB images per object, and crop and resize them into $512\times512$ images used by SAM~\cite{sam} and DUSt3R\cite{wang2023dust3r}.

\begin{figure}
    \centering
    \includegraphics[width=0.99\textwidth]{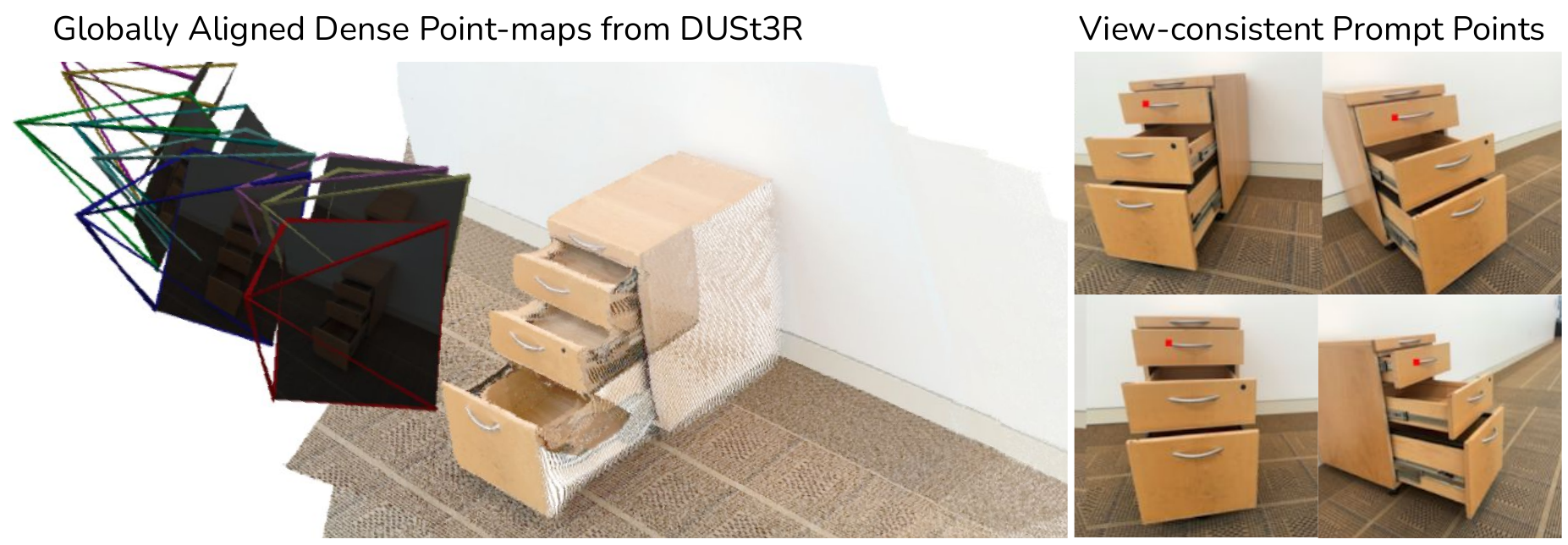}
    \caption{\textbf{3D part segmentation from Pose-free RGB images.} Illustration of how DUSt3R\cite{wang2023dust3r} is used to achieve 3D part segmentation from unstructured RGB images. For each object, we take around 10 pose-free RGB images as input to the pre-trained DUSt3R\cite{wang2023dust3r} model, which outputs a set of globally-aligned 2D-to-3D dense point-maps, i.e.,every 2D pixel on each image is matched to a point in 3D. This correspondence enables cross-view pixel matching via finding nearest-neighbor in 3D space. We can therefore sample view-consistent 2D points for prompting our fine-tuned SAM model, and the resulting segmented masks are grouped into 3D part segmentation. 
    }
    \label{fig:dust3r}
\end{figure}

\textbf{Part Segmentation from Unstructured RGB images.}
\cref{fig:dust3r} visualizes the DUSt3R model output on an example object in: notably, the model predicts dense point-maps on the object's surface area that can be globally-aligned into a object point cloud; but the 3D points are less accurate on the partially occluded areas, such as the inside of the drawer. This is likely due to these areas are less common in the model's pre-training dataset. Also notice that, because we sample each 2D point from one RGB image first and uses nearest neighbor in the predicted 3D point-map to find its matching 2D point in another image, it might find a wrong match if the  point is occluded and not visible in the other image. We address this by manually setting a distance threshold, and decide a match cannot be found if its 3D point's distance to the nearest neighbor is above set threshold. 

\section{Additional Results on Animating Scanned Real World Objects}

\begin{figure} [t]
    \centering
    \includegraphics[width=0.9\linewidth]{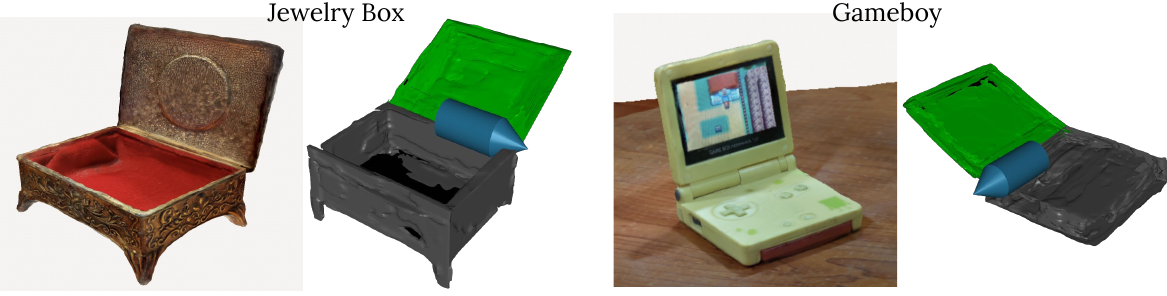}
    \caption{We demonstrate that \ours~can be used for labeling and animating real world objects. We evaluate \ours~on scanned real objects from Polycam\cite{Polycam2024} and export the resulting mesh and joints in MuJoCo~\cite{mujoco}. Blue arrows indicate the simulated joint axis and position; mesh corresponding to the moving part is colored in green.}
    \label{fig:polycam} \vspace{-5mm}
\end{figure}

In addition to object reconstruction from raw RGB images, we show \ours~can also be used to animate scanned objects. We use real world scanned object meshes uploaded by users of the Polycam~\cite{Polycam2024} App, and use our Blender rendering pipeline to render RGB-D images. We evaluate our image segmentation, shape completion, and code generation models on these images, and demonstrate only the qualitative results due to the lack of ground-truth data. We execute the final model output code to show the objects can be simulated in MuJoCo\cite{mujoco}. See~\cref{fig:polycam} for visualizations. These real world objects feature complex visual appearance -- outside our SAM fine-tuning distribution --, but \ours~is still able to successfully segment parts and predict reasonable joint positions and rotations. 

%%%%%%%%%%%%%%%%%%%%%%%%%%%%%%%%%%%%%%%%%%%%%%%%%%%%%%%%%%%%
\end{document}